\documentclass[letterpaper, 10 pt, conference]{ieeeconf}  

\IEEEoverridecommandlockouts                              
\usepackage[utf8]{inputenc}
\usepackage[T1]{fontenc}
\usepackage{gensymb}
\usepackage{amsmath}
\usepackage{graphicx}
\usepackage{subfig}
\usepackage{siunitx}
\usepackage{textcomp} 
\usepackage[linesnumbered,ruled,vlined]{algorithm2e}
\usepackage{xcolor}
\usepackage[normalem]{ulem}

\newcommand{\Ali}{\textcolor{black}}
\newcommand{\Mohsen}{\textcolor{black}}

\providecommand{\keyword}[1]
{
  \small	
  \textbf{\textit{Keywords---}}
}

\title{\LARGE \bf{Robust Environment Perception for Automated Driving: A Unified Learning Pipeline for Visual-Infrared Object Detection}}

\author{Mohsen Vadidar$^{1}$, Ali Kariminezhad$^{1}$, Christian Mayr$^{1}$, Laurent Kloeker$^{2}$ and Lutz Eckstein$^{2}$
\thanks{$^{1}$The authors are with the Elektronische Fahrwerksysteme GmbH, 85080 Gaimersheim, Germany
        {\tt\small \{mohsen.vadidar, ali.kariminezhad, christian5.mayr\}@efs-auto.de}
}
\thanks{$^{2}$The authors are with the research area Vehicle Intelligence \& Automated Driving, Institute for Automotive Engineering, RWTH Aachen University, 52074 Aachen, Germany
        {\tt\small \{laurent.kloeker, lutz.eckstein\}@ika.rwth-aachen.de}}
}

\begin{document}

\maketitle
\thispagestyle{empty}
\pagestyle{empty}

\begin{abstract}
The RGB complementary metal-oxide-semiconductor (CMOS) sensor works within the visible light spectrum. Therefore it is very sensitive to environmental light conditions. On the contrary, a long-wave infrared (LWIR) sensor operating in 8-14 \textmu m spectral band, functions independent of visible light.

In this paper, we exploit both visual and thermal perception units for robust object detection purposes. After delicate synchronization and (cross-) labeling of the FLIR \cite{FLIR} dataset, this multi-modal perception data passes through a convolutional neural network (CNN) to detect three critical objects on the road, namely pedestrians, bicycles, and cars. After evaluation of RGB and infrared (thermal and infrared are often used interchangeably) sensors separately, various network structures are compared to fuse the data at the feature level effectively. Our RGB-thermal (RGBT) fusion network, which takes advantage of a novel entropy-block attention module (EBAM), outperforms the \Mohsen{state-of-the-art network \cite{GAFF} by 10\%} with 82.9\% mAP.
\end{abstract}


\section{INTRODUCTION}
\label{sec:introduction}
A statistical projection of traffic fatalities \Mohsen{in the United States} for the first half of
2021 shows that an estimated 20,160 people died in motor vehicle traffic crashes. This represents an increase of about 18.4 percent as compared to 17,020 fatalities that were reported in the first half of 2020 \cite{Accidents}. Looking at the fatal accidents of 2019 based on the time, one can see that there are 1,000 more fatal accidents during the night-time compared to the day-time \cite{TOD}. Given less average traffic during the night-time, the importance of visibility in dark is inevitable.

The number of publications on RGB-IR sensor fusion for multi-spectral object detection in the automotive sector has increased within the past two years. However, the lack of data in this research area is still noticeable. There are two main sources of data, namely FLIR thermal dataset \cite{FLIR} and KAIST multi-spectral pedestrian detection benchmark \cite{KAIST}, which provide a dataset containing IR and RGB pair images. FLIR mainly provides three classes car, pedestrian, and bicycle, whereas KAIST only contains pedestrians.

\begin{figure}[t]
\centering
\setlength{\lineskip}{\medskipamount}
\subfloat[RGB Frame]{\includegraphics[width=0.48\linewidth]{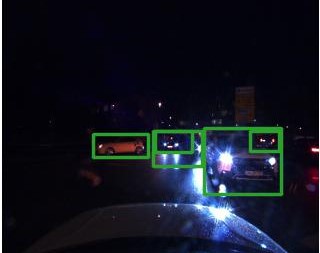}}\hfill
\subfloat[IR Frame]{\includegraphics[width=0.48\linewidth]{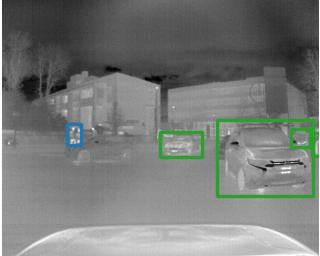}}\hfill
  \caption{A sample of IR and RGB frames in heavy rain from the same scene: one can notice that the pedestrians are not detected on the RGB frame and the cold vehicle coming out of a parking lot is not recognized on the IR frame. Realizing the complementary data in this multispectral setup motivated us to fuse the information.}
\end{figure}

Nevertheless, the FLIR dataset comes only with IR labels. That introduces the first challenge to researchers. Previously published papers \cite{GAFF}, \cite{CaT} and \cite{CFaR} have made various objections to the dataset. For instance, the usage of different camera resolutions at multiple instances ranging from 0.3 to 3.1 MP, misalignment of resolution/aspect ratio, a vast difference between the field of views and resolutions between the sensors, and no possibility to align the RGB and IR frames without having a correlation matrix. These issues currently made fusing the data on this dataset an unsolved problem.

Within this work, we address the FLIR dataset challenges and propose a solution to properly utilize it. Further, we introduce a CNN architecture to fuse RGB-IR data. The proposed network outperforms the state-of-the-art networks for multi-spectral object detection purposes \cite{GAFF}. In section \ref{sec:related_work}, the latest advancements and publications on RGB-IR sensor fusion networks will be presented. Later in section \ref{sec:dataset_and_pre-processing}, a cross-labeling algorithm and pre-processing pipeline will be introduced to provide suitable labels for RGB frames. In section \ref{sec:network_structures}, we start with an RGB network and compare its detection performance with a monospectral IR network. The best monospectral detector will be considered as our baseline. Afterward, we proceed with the simplest form of fusion and develop our proposed RGBT network step by step. Finally, sections \ref{sec:results_and_analysis} and \ref{sec:conclusion} are dedicated to results, analysis and conclusion of this work.



\section{Related Works}
\label{sec:related_work}
In July 2020 Ravi Yadav et al. attempted fusing color and thermal images to detect objects for self-driving applications \cite{CaT}. They have used VGG16 \cite{VGG} and Faster-RCNN \cite{FasterRCNN} as their encoder and detector, respectively. They achieved a miss rate of 29\% on the KAIST benchmark. However, they could not perform well on the FLIR benchmark due to its complications.
Later that month, Chaitanya Devaguptapu et al. \cite{BfA} have used the GAN framework to use the FLIR dataset and produce RGB images from the IR frames. The idea was to borrow knowledge from the data-rich pre-trained RGB domain and use it on IR images.
Heng Zhang et. al. introduced cyclic fuse and refine blocks for object detection \cite{CFaR} in September 2020. The FSDD (Feature Fusion Single Shot Multibox Detector) network \cite{FSDD} is used as their object detector and two VGG16 backbones are implemented to work independently from each other for each sensor. No information is given regarding the number of parameters or complexity of the network. However, looking at the designed cyclic module, the number of its occurrences, and considering the used VGG16, which is one of the most expensive feature extractors with 138 million parameters, one can realize that this network has more than 276 million parameters. In this work, a so-called "well-aligned" version \cite{CFaR} of FLIR is released. However, as mentioned in the paper, the misalignment problem made exploiting FLIR dataset for multispectral object detection impractical. Therefore, the pre-processing part, could not tackle all the issues from the dataset. We will have a look at a failed attempt in the next section. Nevertheless, Cyclic Fuse-and-Refine Network (CFR\_3) could reduce the miss rate on the KAIST dataset and improve the mAP on the FLIR benchmark to 72.39\%.
In November 2020 A. Sai Charan et al. attempted to fuse IR and RGB data by giving weight to each sensor using attention mechanism \cite{AbiFN}.
The attention module is placed in every residual block \cite{resnet101} after the RELU activation function and first batch normalization. They have tried both the late and mid-fusion approaches. Late fusion in this case is done by pooling features of the thermal and RGB camera separately from their respective region proposals and then concatenating them together. Comparing the performance of the two mid and late fusions, mid-fusion has shown a better result with 62.12\% mAP.
In January 2021 Heng Zhang et. al. proposed guided attentive feature fusion (GAFF) \cite{GAFF}. This paper claims to be the first work that is considering multi-spectral feature fusion as a sub-task in network optimization. While bringing accuracy gains, GAFF has a lower computational cost compared to common addition or concatenation methods. They reduced the miss rate on the KAIST dataset by 2\% and achieved an mAP of 72.9\% on the FLIR dataset. To the best of our knowledge, this is the highest mAP achieved using the FLIR dataset. Thus, this work will be taken as the state-of-the-art network on this benchmark. ZHANG et. al. in a comprehensive review studied different methods and levels of fusion between visible and IR cameras for the task of tracking \cite{ZHANG2020166}. Since the feature extraction process is the same in tracking and object detection, this study can provide a good insight into feature-level fusion using deep learning-based methods.
 

\section{Dataset and Data Pre-processing}
\label{sec:dataset_and_pre-processing}
This section is an important part of our work, without which the rest of our research would not be possible.

\subsection{Dataset}
Table \ref{tab:datasets} summarizes open-source datasets including IR frames. Most of the datasets are concerned with pedestrian class only. Those which cover a wider range of objects like the Dense dataset \cite{Dense} confront other downsides such as non-availability of RGB-IR data pairs, annotation issues, strong misalignment between IR and RGB fields of view (FOV), and lack of extrinsic calibration information (e.g. correlation matrix).

\begin{table}[b]
    \caption{Available Datasets including IR Frames \Mohsen{(NG stands for not given)} }
    \label{tab:datasets}
    \begin{center}
    \begin{tabular}{|l|l|l|l|}
    \hline
        \bf{Name} & \bf{Year} & \bf{Size \Mohsen{(Frames)}} & \bf{Downsides} \\ \hline
        FLIR & 2020 & \Mohsen{13813} & IR label only \\\hline
        Dense & 2020 & 121500 & Narrow IR FOV \\\hline
        ZUT-FIR & 2020 & \Mohsen{NG} & IR data available only \\\hline
        SCUT-CV & 2019 & \Mohsen{211011} & IR data available only\\
        &  &  & Pedestrian class only \\ \hline
        CAMEL & 2018 & \Mohsen{NG} & Not on-board frames \\\hline
        Multispec. & 2017 & 7512 & Different Ref. Sys.\\\hline
        CVC-14 & 2016 & 3695 & Gray RGB\\
        &  &  & Pedestrian class only  \\ \hline
        KAIST & 2015 & \Mohsen{95000} & Crowd label as human \\
        &  &  & Pedestrian class only\\ \hline
        CVC-09 & 2014 & 5990 & Pedestrian class only \\ \hline
    \end{tabular}
    \end{center}
\end{table}

\subsection{Data Pre-processing}
To include the Vulnerable Road Users (VRU) as well as motorized users in our study, we have decided to work with FLIR dataset \cite{FLIR}. The non-availability of a dataset for fusing the IR and RGB frames, led us to put a great deal of effort into developing a reliable cross-labeling algorithm, additional manual labeling of the missing objects, and checking the data one by one, to make sure the entire dataset is reliable and usable.

Sensor data fusion between an IR and an RGB camera was a challenge to researchers since there is no open-source dataset providing synchronized and labeled RGB-IR image pairs. Given that the ground truth in FLIR dataset is provided for IR frames only, we introduce a method to cross-label the RGB frames using IR annotations semi-automatically.

\begin{figure}[t]
    \centering
    \includegraphics[width=1\linewidth]{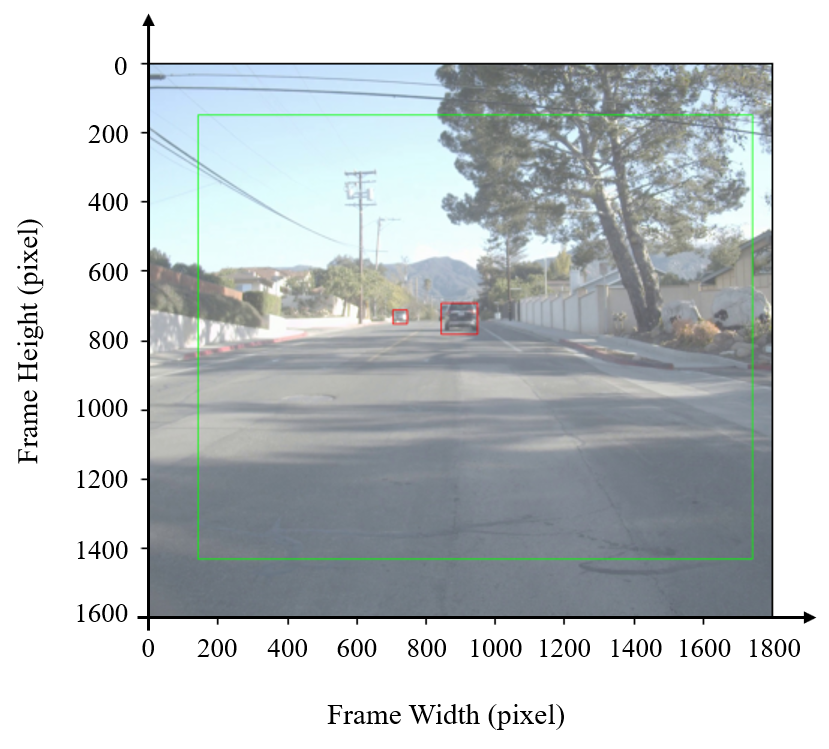}
    \caption{IR-RGB overlay: Union of the filed of views in green and the transferred labels from IR to RGB frame in red box}
    \label{fig:CL_CC}
\end{figure}

\begin{figure}[t]
\centering
\setlength{\lineskip}{\medskipamount}
\subfloat["well-aligned" version \label{fig:RGB}]{\includegraphics[width=0.48\linewidth]{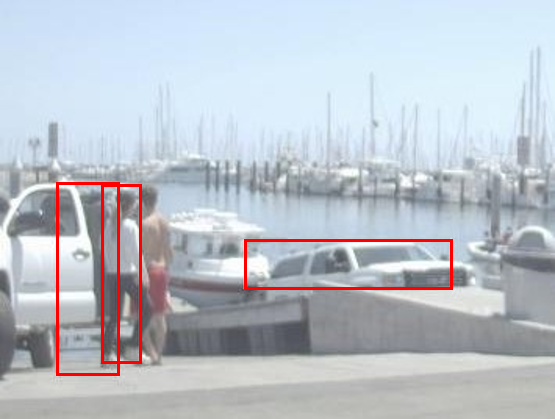}}\hfill
\subfloat[our version \label{fig:IR}]{\includegraphics[width=0.48\linewidth]{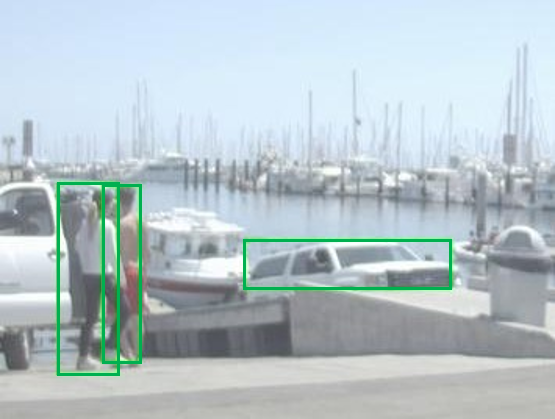}}\hfill
  \caption{Comparison of two cross-labeling methods: the left image represents the previous attempt \cite{CFaR} to process the FLIR dataset \cite{FLIR} and transfer the labels from IR to RGB coordinate system. The right picture illustrates the result of our proposed cross-labeling algorithm.}
    \label{fig:CCC}
\end{figure}

Algorithm \ref{alg:CC} demonstrates our approach for pre-processing the data. We first convert the RGB frames to gray-scale images and extract the edges. IR and RGB frames are matrices of size $640\times512$ and $1800\times1600$, respectively. Thus, we introduce a scaling-parameter to resize the IR frame gradually in a loop, until the IR extracted edges fall on RGB edges. The cross-correlation value between the two matrices is a good metric to understand the accuracy of the alignment. The highest value is obtained, when the edges of two frames are completely aligned. Since the two frames are different in size, the cross-correlation between them takes place in a sliding-window manner. As figure \ref{fig:CL_CC} illustrates, vertical and horizontal offsets exist between the RGB and IR frames. Figure \ref{fig:CCC} represents a comparison between our result and the "well-aligned" version \cite{CFaR}. Ultimately, we utilize 6924, 1982, and 3960 pair images for training, validation, and test, respectively.

\begin{algorithm}[!htbp]
\SetAlgoLined
\KwResult{X\_Offset, Y\_Offset, Scaling\_Factor}
 load both frames \\
 Smooth both frames by a Gaussian filter \\
 Detect the edges using Canny edge detector \\
 Use binary threshold to set the edge pixels to one and all the other pixels to zero\\

Scaling\_Factor = 2 \# A constant to resize the IR frame\\
 \While{Width\_IR < Width\_RGB}{
 Width\_IR\_new := Width\_IR $\times$ Scaling\_Factor\\
 Resize IR binary frame with the Width\_IR\_new and the corresponding Height\_IR\_new\\
 Cross correlate (CC) the IR and RGB binary frames in a sliding window manner and save the pixel coordinate of the maximum value from the correlation matrix\\
  
  \If{CC\_Value > Best\_CC\_Value}{
   Update Best\_CC\_Value, Pixel\_coordinates, Best\_Scaling\_Factor\\
   }
  Scaling\_Factor += 0.001\\
 }
 Best\_X\_Offset, Best\_Y\_Offset := Pixel\_coordinates\\
 return Best\_X\_Offset, Best\_Y\_Offset, Best\_Scaling\_Factor\\
 \caption{Cross Labeling Algorithm}
 \label{alg:CC}
\end{algorithm}

\section{Network Structures}
\label{sec:network_structures}
One of the most accurate networks on MSCOCO benchmark \cite{COCO}, namely Scaled-Yolov4 \cite{Wang_2021_CVPR} is selected as a starting point for this work. The performance of each sensor will be evaluated separately and the best monospectral network will be chosen as our baseline. The CSP (Cross-Stage-Partial-Connections) \cite{CSP} version of the network is used for both sensors. That is possible since the IR sensor is delivering image-like data and the encoder is very good at picking up the features and adapting itself to the IR sensors' data texture (i.e. domain adaptation).

\begin{figure*}[!htbp]
    \centering
    \includegraphics[width=1\linewidth]{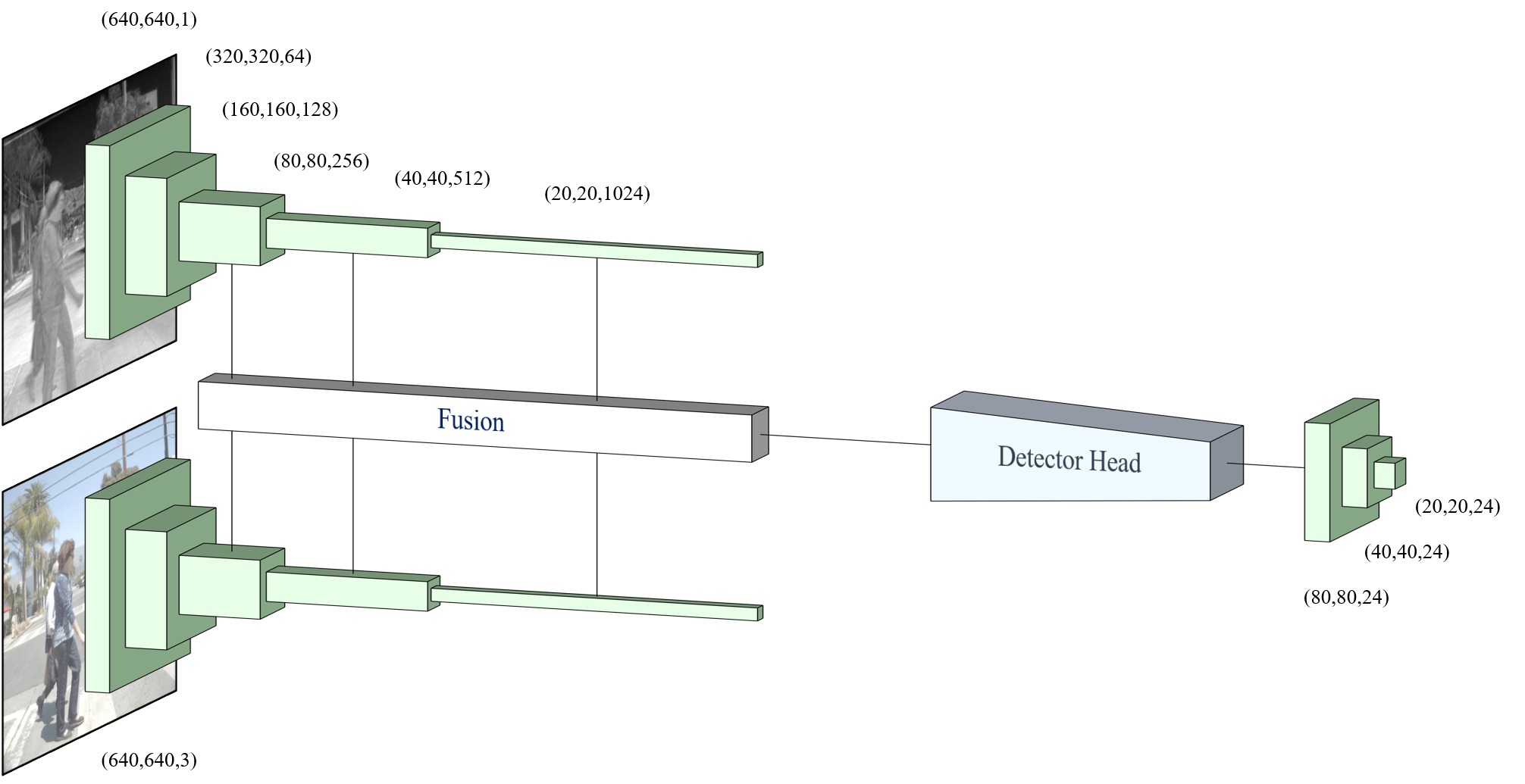}
    \caption{RGBT Network Structure: Two identical encoders are used for feature extraction from both RGB and IR frames. The feature-level fusion block receives features in three scales from each encoder and fuses them. Then, the data is passed to the detector, which is the same as CSP version of Scaled\_Yolov4. At last, three tensor outputs refer to three scales. Elements of the fusion block are demonstrated in figure \ref{fig:Fusion_Blocks}.}
    \label{fig:RGBT}
\end{figure*}

\begin{figure}[!htbp]
    \centering
    \includegraphics[width=1.0\linewidth]{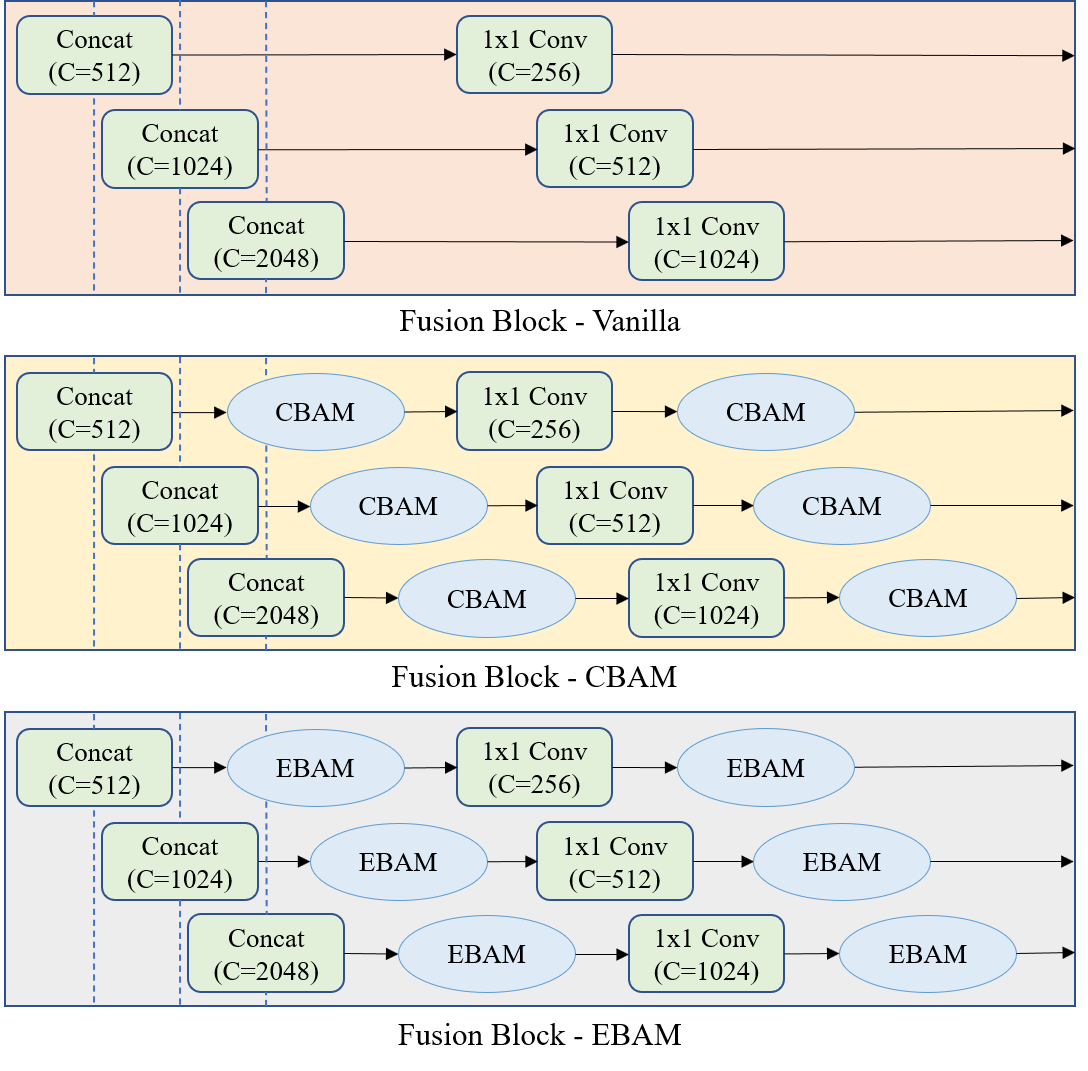}
    \caption{Upper block: Vanilla fusion is referred to a simplistic approach to fuse the data by just concatenating the extracted features in each scale and selecting the most valuable information by a convolution layer. Middle block: CBAM is added as an attention module. Lower block: EBAM is designed to select informative features more efficiently.}
    \label{fig:Fusion_Blocks}
\end{figure}

To take the most of each monospectral backbone, we start with our vanilla fusion. The designed network fuses the feature maps across scales as shown in figures \ref{fig:RGBT} and \ref{fig:Fusion_Blocks}. The smallest scale is designed for large objects and the biggest one detects the smaller objects. After concatenation, channels are halved using $1\times 1$ convolution layers to prepare the feature maps for the detector head.


The most crucial part of fusing visible and IR sensors is finding a way to weight each backbone, depending on how informative the extracted features are. As each sensor receives a different wavelength, we want to ensure that complementary data from both spectrums increase the chance of detecting an object as well as improving the robustness of the network. The goal is to detect the objects, even in lack of strong emissions within visible-light frequency. To do that, we integrate two Convolutional Block Attention Modules (CBAM) \cite{Woo_2018_ECCV}, such that more weight is given to more informative features depending on the input data (see figure \ref{fig:Fusion_Blocks}).

~\Ali{In what follows, we will discuss our proposed entropy-block attention module (EBAM) and how it is a better substitute for the conventional CBAM \cite{Woo_2018_ECCV}. In this regard, we will proceed with channel attention, which can further be generalized to spatial attention. Suppose, average and max pooling for feature map (channel) $k$ are represented by $a_k$ and $m_k$, as:
\begin{align}
a_k&=\frac{1}{MN}\sum_{i}^{M}\sum_{j}^{N}X^{<k>}_{ij},\\
m_k&=\max_{i,j}\{X^{<k>}_{ij}\},
\end{align}
respectively, where $\mathbf{X}^{<k>}$ is the matrix of $k$th feature map and $X^{<k>}_{ij}$ is the $k$th feature map grid value for the grid located in $i$th row and $j$th column. Further, we define the vector of average and max poolings as
\begin{align}
\mathbf{a}&=[a_1,\cdots,a_k,\cdots,a_{K}]^T,\\
\mathbf{m}&=[m_1,\cdots,m_k,\cdots,m_{K}]^T.
\end{align}
According to the CBAM, the vectors $\mathbf{a}$ and $\mathbf{m}$ pass through a shared multi-layer perceptron, for which the weights are optimized in the backpropagation. Here, we assume a single layer for convenience in further mathematical presentation. This can later be generalized to multi-layer networks. Thus, the channel attention portion of the CBAM with a single-layer perceptron delivers the following weight vector:
\begin{align}
    \mathbf{c}_{\textrm{CBAM}}=\mathbf{W}(\mathbf{m}+\mathbf{a})+\mathbf{b},
\end{align}
where the matrix of shared weights and the bias are represented by $\mathbf{W}\in R^{K\times K}$ and $\mathbf{b}\in R^{K}$, respectively. Therefore, the attention weight corresponding to the $k$th feature map can readily be formulated as
\begin{align}
    c^{<k>}_{\textrm{CBAM}}=\mathbf{w}_k(\mathbf{m}+\mathbf{a})+b_k,
\end{align}
where $\mathbf{w}_k$ is the $k$th row of matrix $\mathbf{W}$. Interestingly, one could observe that the average and max poolings are equally weighted due to the shared layers among them. \Mohsen{Consequently, the contribution of the average and max poolings are not captured and only the sum of the values plays a role.} For instance, assume a particular feature map consists of multiple very high values, while the majority of the grids contain low values. This way, the average pooling downgrades the importance of that particular feature map. However, due to the shared weights, this phenomenon can not be remedied in the backpropagation by weight optimization.
This motivates us to develop a more robust and flexible attention mechanism as a replacement for CBAM.\\
Uncertainty of a data-driven feature is a valuable metric delivering a notion for assessing how much extra information can still be mined from that feature. 
\begin{align}
H_k=-\sum_{i=1}^{M}\sum_{j=1}^{N}P(X^{<k>}_{ij})\log P(X^{<k>}_{ij}),    
\end{align}
where $P(X^{<k>}_{ij})$ are the probabilities delivered by the softmax operation as,
\begin{align}
P(X^{<k>}_{ij})  = \frac{e^{X^{<k>}_{ij}}}{\sum_{i=1}^M \sum_{j=1}^N e^{X^{<k>}_{ij}}}.
\label{eq:H}
\end{align}
Notice that, equal probabilities maximizes the entropy.
\begin{align}
P(X^{<k>}_{ij})=\frac{1}{MN},\quad\forall i,j.    
\end{align}
It is important to discern that, the max entropy is reached if the max pooling equals the average pooling of the feature map $k$, i.e., $a_k=m_k$, which holds if:
\begin{align}
 X^{<k>}_{ij}=\frac{1}{\sum_{i}^{M}\sum_{j}^{N}X^{<k>}_{ij}},\forall i,j.
 \label{eq:lower-bound}
\end{align}
This reflects the fact that in an extreme case the sum of max and average, i.e., $m_k+a_k$, corresponds with the entropy upper-bound, i.e., $H^{\textrm{max}}_k$. This extreme case, i.e., $a_k=m_k$ can not be differentiated by CBAM, since it considers only the sum. However, entropy captures the interaction between max and average as follows:
\begin{align}
    \lim_{m_k-a_k\rightarrow 0}H_k&=H^\textrm{max}_k,\\
    \lim_{m_k-a_k\rightarrow \infty}H_k&=0.
\end{align}
That means, as the max pool gets closer to the average pool, the entropy maximizes and as it gets higher value than the average pool, the entropy approaches zero.
It is important to mention that, the entropy of a random process does not reduce except by providing information. Furthermore, notice that more data does not necessarily mean more information. This observation motivated us to design a network, which focuses on informative features in the fusion process. This focus is represented by strongly weighting the highly-informative feature maps. We call this new attention mechanism entropy-block attention module (EBAM), for which the $k$th channel attention weight is given by:
\begin{align}
    c^{<k>}_{\textrm{EBAM}}=\mathbf{v}_k\mathbf{h}+b_k,
\end{align}
where $\mathbf{h}=[H_1,\cdots,H_k,\cdots,H_{K}]^T$. Note that $\mathbf{v}_k$ is the channel attention weight vector designated for the $k$th feature map optimized in the backpropagation. In what follows we generalize the EBAM to capture both channel and spatial attention.\\
Let's define the intermediate feature map $ F \in R ^ {C\times H\times W}$ as a tensor consisting of $\mathbf{X}^{<k>},\ \forall k$. That means, $F=[\mathbf{X}^{<1>},\cdots,\mathbf{X}^{<k>},\cdots,\mathbf{X}^{<K>}]$. This feature map tensor is the input to our attention module, from which attention weights are derived. Hence, the generalized self-attention can be written as:
\begin{equation} 
W(F) = \alpha(F) \:\otimes\: F,
\end{equation}
where $\otimes$ denotes element-wise multiplication. $C, H$ and $W$ refer to channel, height and width, respectively and $\alpha(F):R ^ {C \times H \times W}\rightarrow R ^ {C \times H \times W}$ is an attention function of the feature map $F$.}

\begin{figure}[ht]
    \centering
    \includegraphics[width=1\linewidth]{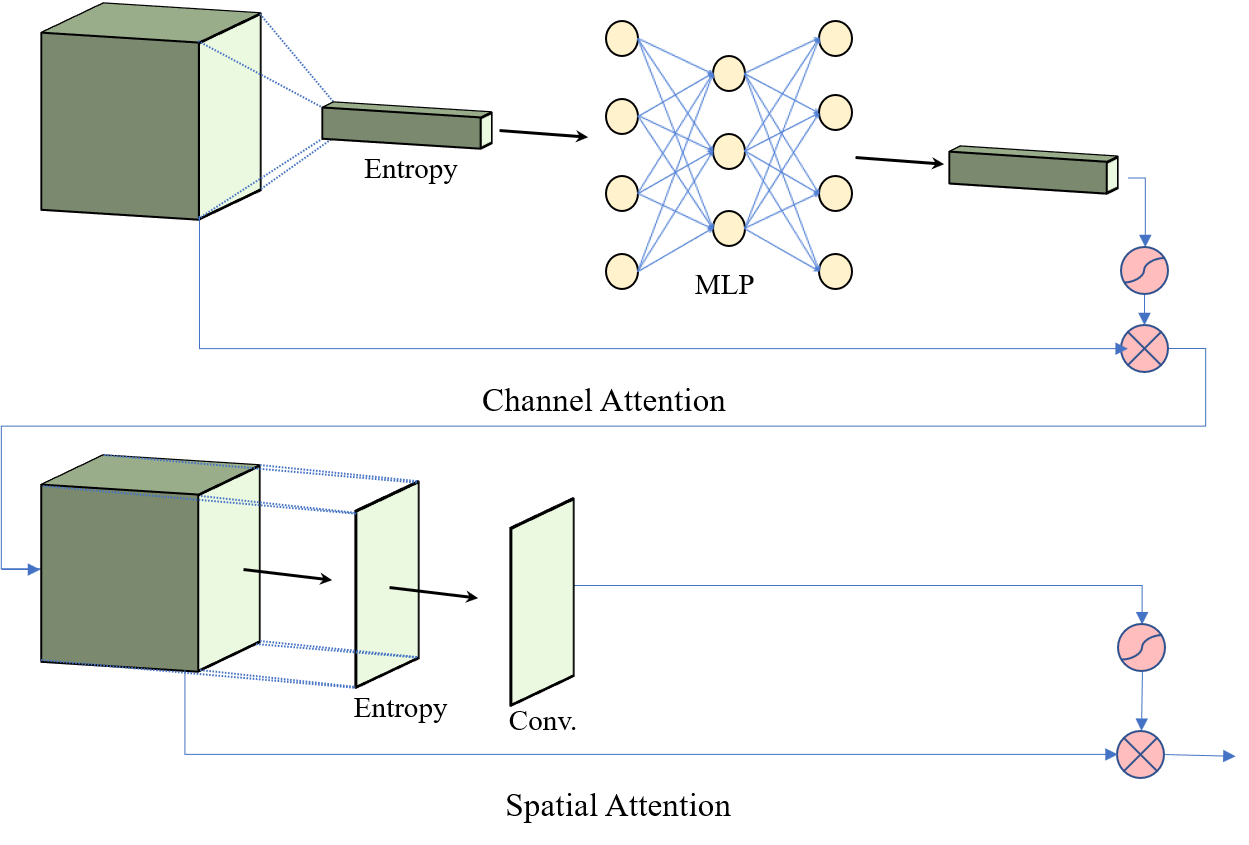}
    \caption{EBAM: detailed illustration of entropy-block attention module, which contains channel and spatial domains. MLP refers to multilayer perceptron.}
    \label{fig:EBAM}
\end{figure}

EBAM sequentially infers a 1D channel attention map $A_{c} \in R ^ {C \times 1 \times 1} $ and a 2D spatial attention $A_{s} \in R ^ {1\times W \times H} $ as illustrated in figure \ref{fig:EBAM}. The overall attention process can be summarized as:
\begin{equation}
    \begin{aligned}
        F^\prime = A_{c}(F) \otimes F, \\
        F^{\prime\prime} = A_{s}(F^\prime) \otimes F^\prime,
    \end{aligned}
\end{equation}

In the channel attention by giving more weight to more uncertain feature maps (assuming each feature map is represented by a random variable), the network is signaled to extract more information by prioritizing the gradient direction in backpropagation. Therefore, the network is forced to gain more information (i.e. kernel activation) from different kernels. As a matter of fact, the proposed attention module operating only in the channel domain, showed a relatively better performance than a CBAM including both spatial and channel attention. One can calculate the explained channel attention as follows:
\begin{equation}
    \begin{aligned}
        A_c(F) &= \sigma(MLP(H(F))), \\
               &= \sigma(W_0(F^C_H)),
    \end{aligned}
\end{equation}

where $\sigma$ denotes the sigmoid function, $H$ is the entropy matrix, $F$ stands for feature map and $W_0$ represents the MLP weights.

The more the entropy of a random process, the more uncertain we are about the outcome. In the case of the attention module in the spatial domain, each grid pattern can contribute to detecting an object. As the training process goes on, the grids, which refer to an object will have lower entropy (i.e. lower uncertainty). Hence, if we use the entropy values directly, more weight will be given to grids with high entropy, which would be the background information in this case. Thus, more weights are intended to be allocated for the grids with less entropy values through the feature maps.

Therefore, the entropy-based spatial attention can be denoted as follows:
\begin{equation}
    \begin{aligned}
        A_s(F^\prime) &= \sigma(Conv.(1 - \dfrac{H_{ij}}{\max_{\{i, j\}} H_{ij}})), \\
                      &= \sigma(W_1(F^S_H)),
    \end{aligned}
\end{equation}

The networks are trained to the point, where an overfitting effect appeared. The hyperparameters are kept unchanged as the original Scaled-Yolov4 \cite{Wang_2021_CVPR} implementation.



\section{Results and Analysis}
\label{sec:results_and_analysis}
In this chapter, the results of our pre-processing, baseline, and fusion networks are presented. The processed data and the code can be found in the following GitHub link: https://github.com/SamVadidar/RGBT.
\subsection{Data Pre-processing}
Exploiting our proposed pre-processing step (i.e. algorithm 1), not only the accuracy of the bounding boxes is improved, but more data (i.e. 13,848 images) could be recovered from the original dataset. That means 34.6\% more data with respect to what Heng Zhang's version \cite{CFaR} could use in training. Thus, a dataset is produced, where the labels can be referred to both frames, which are $640\times512$ pixels. After running the pre-processing algorithm on the original dataset, one receives 6924, 1982, and 3960 paired images for train, validation, and test sets, respectively. Since the intention is to compare a CMOS stand-alone sensor with a CMOS-IR sensor pair, the dataset is split into night and day subsets. Therefore, one can test and evaluate a trained network based on these two scenarios.

\subsection{Studied Networks}
To reduce the computational costs, all the network trainings and their comparisons will be done using $320\times 320$ image size and only the final network will be fine-tuned with the original images ($640\times 512$ pixels). Generally speaking, car class being the largest object among the three, shows the best mAP. On the contrary, the person class being the smallest in size is the most challenging object to detect.

\begin{table}[b]
    \caption{Summary of studied networks:
    Fusion+H\_C refers to a fusion network with entropy-block attention modules in the channel domain only. RGBT contains the full network structure as discussed in chapter \ref{sec:network_structures}. \Mohsen{CFR \cite{CFaR} and GAFF \cite{GAFF} illustrate the two best available mAPs (NG stands for not given) on the FLIR dataset \cite{FLIR}. RGBT* is trained and tested using previously published aligned version \cite{CFaR} of the FLIR dataset.}}
    \label{tab:comparison}
    \begin{center}
    \begin{tabular}{|l|l|l|l|l|l|}
    \hline
        Network & \multicolumn{4}{c|}{mAP@.5} & Param. \\ \hline
        & Person & Bicycle & Car & Overall &    \\ \hline
        RGB Baseline & 39.6\% & 50.4\% & 79.4\% & 56.6\% & 52.5 M \\ \hline
        IR Baseline & 49.6\% & 54.9\% & 84.4\% & 63.0\% &  52.5 M \\ \hline
        Vanilla Fusion & 56.9\% & 56.7\% & 82.0\% & 65.2\% &  81.8 M \\ \hline
        Fusion+CBAM & 57.6\% & 60.5\% & 83.6\% & 67.2\% &  82.7 M \\ \hline
        Fusion+H\_C & 62.6\% & 65.9\% & 86.0\% & 71.5\% &  82.7 M \\ \hline
        RGBT & 63.7\% & 67.1\% & 86.4\% & 72.4\% & 82.7 M \\ \hline
        CFR(640) & 74.4\% & 57.7\% & 84.9\% & 72.3\% & $>$276 M \\ \hline
        \Mohsen{GAFF(640)} & NG & NG & NG & 72.9\% & NG \\ \hline
        \Mohsen{RGBT*(640)} & \textbf{85.3\%} & 64.0\% & 88.6\% & 79.3\% & 82.7 M \\ \hline
        RGBT(640) & 80.1\% & \textbf{76.7\%} & \textbf{91.8\%} & \textbf{82.9\%} & 82.7 M\\ \hline
    \end{tabular}
    \end{center}
\end{table}

A very interesting finding is that the infrared camera outperforms the CMOS sensor even as a stand-alone sensor. Its performance in table \ref{tab:comparison} shows a 6.4\% overall improvement in the mAP metric using the same training pipeline, same network, and an equal number of epochs. One can observe the improvement in all classes, especially for the pedestrian class with a 10\% jump in mAP. High contrast in the frames between the objects and surroundings, very high sensor sensitivity of less than 50 mK as well as an operating temperature of -40\textdegree C to +85\textdegree C, delivers a data-rich infrared frame for feature extraction. Later in this chapter, some qualitative and quantitative analyses will be illustrated to support this statement.

The first attempt for fusing the two backbones (i.e. Vanilla Fusion) has shown a 2.2\% improvement in the overall mAP compared to the IR network (See table \ref{tab:comparison}). The boldest improvement among the classes belongs to the pedestrian class, which has experienced a drastic improvement of 7.3\% with respect to the IR network and 17.3\% to the RGB network considering the mAP metric. Overall 8.6\% mAP rise compared to the RGB network by adding no additional module, motivated us to use the CBAM attention module \cite{Woo_2018_ECCV}.
By the way the CBAM is implemented, each beneficial sensor data is enhanced and downsides are compensated by focusing on the right and reliable features automatically. Therefore, an additional performance gain of 2.2\% compared to the vanilla network is achieved and the bicycle class mAP is improved by 3.8\%. After designing the entropy-block, despite spatial attention absence, the Fusion+H\_C network could outperform the Fusion+CBAM network. One can see 4.3\% mAP improvement overall, and 5.0\% for pedestrian class.

The proposed RGBT Network contains two complete (i.e. channel and spatial) attention modules. The spatial attention added 0.9\% on top of the channel-only attention network and pushed the overall mAP up to 72.4\%. The network shows a noticeable overall mAP improvement to RGB and IR baselines with 15.8\% and 9.4\%, respectively. This shows that a fusion network can have an immense impact on improving the mAP. Since the RGBT network performs better than both IR and RGB object detectors, one can observe the importance of an efficient feature-fusion by selecting the best features from each encoder through the entropy-block attention module. This work could outperform the results of the CFR network \cite{CFaR}, by only utilizing half of the pixel resolution. \Mohsen{We also have trained and tested our proposed network with the same aligned version of the FLIR dataset \cite{CFaR}, which CFR and GAFF used. This helps to differentiate the contribution of our preprocessing method from the proposed network. As the table \ref{tab:comparison} illustrates, 3.6\% mAP improvement is solely achieved by better preprocessing the data. The test-set of the "well-aligned" dataset has almost 46\% less number of pedestrian instances. Fewer test data and relatively easier scenarios might be the underlying reason behind the higher mAP (i.e. 85.3\%) for this class. Considering the number of parameters and what the CFR paper \cite{CFaR}} reports, two VGG-16 networks \cite{VGG} are used as separate feature extractors of each sensor. Hence, knowing that each VGG-16 has around 138 million parameters, one can find out that the number of the parameters in the state-of-the-art network is greater than 276 million. Looking at the table \ref{tab:comparison}, it is demonstrated that the RGBT network could outperform the CFR network by using at least 70\% less number of parameters. Therefore, not only the mAP is improved in this work, but the complexity of the network is reduced drastically. This will cause a considerably faster inference speed compared to the CFR network. \Mohsen{The GAFF, which showed slightly (i.e. 0.6\%) better mAP than CFR network, does not \Ali{report} the per class mAPs and the total number of parameters of the complete network.}
Due to limited resources and long training time, only our proposed network is trained with the original image size. Table \ref{tab:comparison} shows that using the full resolution, one can reach 82.9\% mAP with \Mohsen{10\%} performance gain compared to the state-of-the-art network.

\begin{figure}[t]
\centering
\parbox{8cm}{
\includegraphics[width=1.9cm]{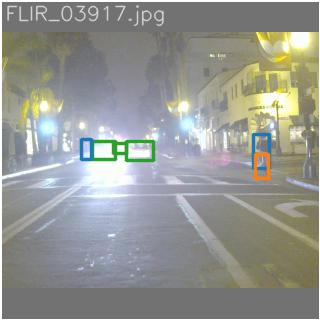}
\includegraphics[width=1.9cm]{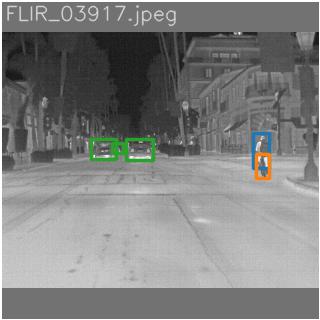}
\includegraphics[width=1.9cm]{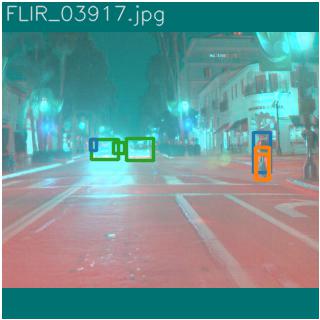}
\includegraphics[width=1.9cm]{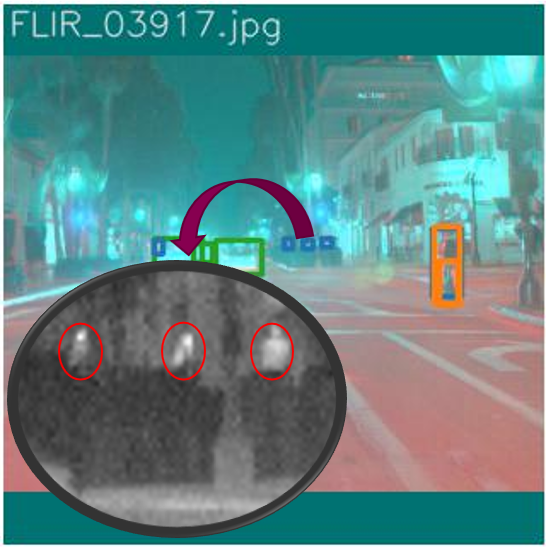}

\label{fig:2figsA}}
\qquad
\begin{minipage}{8cm}
\includegraphics[width=1.9cm]{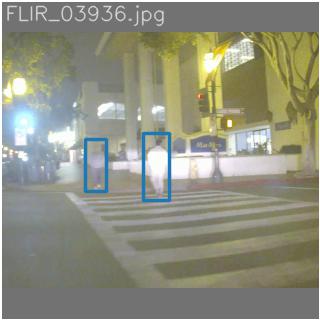}
\includegraphics[width=1.9cm]{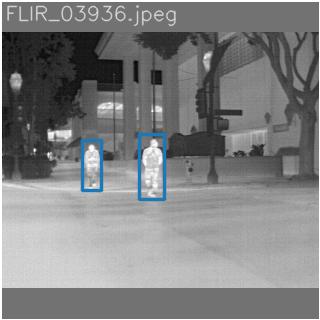}
\includegraphics[width=1.9cm]{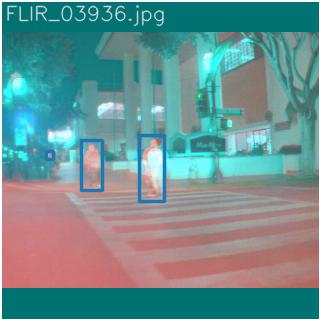}
\includegraphics[width=1.9cm]{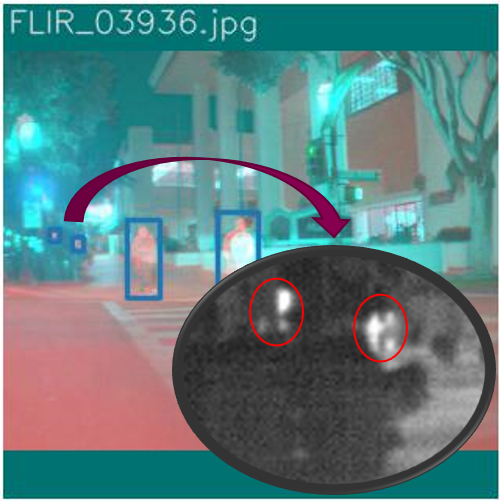}

\includegraphics[width=1.9cm]{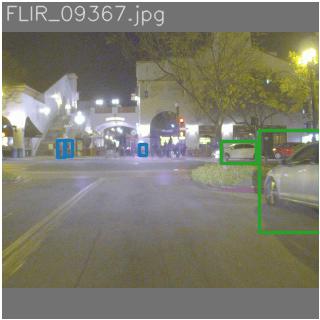}
\includegraphics[width=1.9cm]{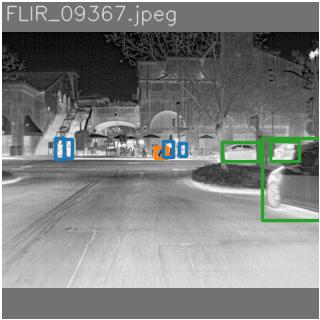}
\includegraphics[width=1.9cm]{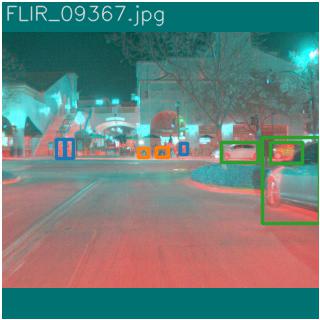}
\includegraphics[width=1.9cm]{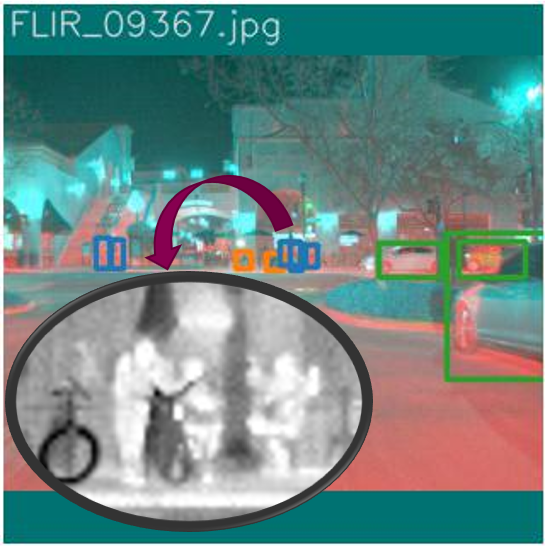}
\caption{Qualitative comparison of three scenarios:
Columns from left to right: RGB, IR, Fusion with CBAM, and Fusion with EBAM. Note that the fusion results are represented using TGB (i.e. thermal, green, blue) instead of RGB colormap.}
\label{fig:2figsB}
\end{minipage}
\end{figure}

The results using the original images are compared to other networks in figure \ref{fig:2figsB}. 10.5\% raise in the mAP due to a higher input resolution, which indicates the importance of the pixel per object value. \Mohsen{Working with FLIR dataset \cite{FLIR}, one can observe that it contains very small and challenging objects, especially in the pedestrian class.} Therefore, when the number of input pixels, which are fed to the network is doubled (i.e. $640 \times 640$), the overall outcome elevates significantly.
Looking at figure \ref{fig:2figsB}, it is to be noted that the detected objects on the sidewalk in the first two scenarios (i.e. first two rows) were not labeled. That means the network is using the extracted features from other instances of the pedestrian class to classify and localize the objects. Since in the training process, detected objects, which do not have a corresponding label are punished by the loss function, the performance of the network in this scene is admirable. The third scenario (i.e. third-row) illustrates a group of heavily occluded pedestrians sitting around a table. The night scenarios are the most challenging scenarios for the RGB cameras and in this case, the target object was neither labeled nor provided a sufficient amount of pixel information. Despite these challenges, the RGBT network having only $320 \times 320$ resolution could detect the pedestrians as well as the bicycles in front of them.




\section{Conclusion \Mohsen{and Outlook}}
\label{sec:conclusion}
In general, cameras are one of the most informative sensors on the vehicle. However, they are highly sensitive to light and weather conditions. Such disturbances affect their performance severely. In order to introduce robustness to the system, the functionality of an infrared camera as a complementary sensor is investigated.
Two groups of researchers \cite{GAFF}, \cite{CaT} and \cite{CFaR} have considered data pre-processing and preparation of FLIR dataset \cite{FLIR} as one of the major challenges and referred to it as an unsolved problem. High-quality data being the core of machine learning approaches motivated us to address this issue with our cross-labeling algorithm. Hence, 34.6\% more data is retrieved with respect to the previous approach \cite{CFaR}.
Considering the trend in recent publications, the rally was to gain a single additional mAP improvement, whereas thanks to the pre-processing pipeline and the entropy-block attention module, the proposed RGBT network could \Mohsen{outperform the state-of-the-art network \cite{GAFF} by 10\% mAP. Undoubtedly, the availability of higher quantity and quality labeled data is essential for further investigation in this domain. In our next step, we are planning to combine our RGB-Radar network with the proposed RGBT network to extract depth information from the detected objects. We also have seen that Radar can improve the mAP considerably. At last, the network will be optimized by TensorRT for our target hardware and run on our demonstrator vehicle to perform real-world tests.}

\addtolength{\textheight}{-11cm}   

\section*{ACKNOWLEDGMENT}

\Mohsen{The authors would like to thank EFS MLOps team, specially Mr. Sebastian Balz for his unceasing support, without which the training and evaluation of the networks would not be possible.}


\bibliographystyle{IEEEtran} 
\bibliography{references}

\end{document}